# Investigation and Assessment of Disorder of Ultrasound B-mode Images

Vidhi Rawat *, Alok jain**, Vibhakar shrimali***,
*Department of Biomedical Engineering
**Department of electronics Engineering
*** Department of Trg. & Technical Education, Govt. of NCT Delhi, Delhi.
Samrat Ashok Technological Institute
Vidisha, India.
Vidhi_pearl@rediffmail.com

*Abstract*— Digital image plays a vital role in the early detection of cancers, such as prostate cancer, breast cancer, lungs cancer, cervical cancer. Ultrasound imaging method is also suitable for early detection of the abnormality of fetus. The accurate detection of region of interest in ultrasound image is crucial. Since the result of reflection, refraction and deflection of ultrasound waves from different types of tissues with different acoustic impedance. Usually, the contrast in ultrasound image is very low and weak edges make the image difficult to identify the fetus region in the ultrasound image. So the analysis of ultrasound image is more challenging one. We try to develop a new algorithmic approach to solve the problem of non clarity and find disorder of it. Generally there is no common enhancement approach for noise reduction. This paper proposes different filtering techniques based on statistical methods for the removal of various noise. The quality of the enhanced images is measured by the statistical quantity measures: Signal-to-Noise Ratio (SNR), Peak Signal-to-Noise Ratio (PSNR), and Root Mean Square Error (RMSE).

I. INTRODUCTION

Ultrasound imaging method is suitable to diagnose and progenies [1]. The accurate detection of region of interest in ultrasound image is crucial. Since the result of reflection, refraction and deflection of ultrasound waves from different types of tissues with different acoustic impedance. Usually, the contrast in ultrasound image is very low and boundary between region of interest and background are fuzzy [2]. fetus region in the ultrasound image is not approachable. So the analysis of ultrasound image is more challenging one .Noise is considered to be any measurement that is not part of the phenomena of interest .Digital image plays a vital role in the early detection of abnormality in the fetus. Processes going on in the production and capture of real signal. Ultrasound is safe, radiation free, faster and cheaper. Ultrasound images themselves will not give a clear view of an affected region. So digital processing can improve the quality of raw ultrasound images. In this work a software tool called Image Processing Tool has been developed by employing the histogram equalization and region growing approach to give a clearer view of the affected regions in the abdomen. Ultrasound is applied for obtaining images of almost the entire range of internal organs, these include the kidney, liver, pancreas, bladder, the fetus during pregnancy. Today, B-mode ultrasound imaging is one of the most frequently used diagnostic tools, not only because of its real time capabilities, as it allows faster and more accurate procedures, but also because there is low risk to the patient and low cost as compared to other image modalities.

A proposed algorithm has been successfully developed to semi-automate noninvasive examination of the fetus. Such a system can be helpful in reducing costs, minimizing exposure of the fetus to ultrasonic radiation, and providing a uniform examination and interpretation of the results.

II. OBSTETRICAL ULTRASOUND IMAGING

Normally ultrasound images will contain a mixture of different types of noises. Removal of noises is crucial since ultrasound images themselves will not give a clear view of an affected region. It is easy to remove the noises if the images are digitized. Then it is possible to develop software that suits to remove a type of noise. After removing the noises, clear view of affected regions can be pinpointed. In the ideal case of a continuous probability distribution of the gray levels, histogram equalization produces a uniform histogram. Ultrasound exams do not use ionizing radiation (as used in x-rays). Because ultrasound images are captured in real-time. Ultrasound imaging is a noninvasive medical test that helps physicians diagnose and treat medical conditions. Obstetrical ultrasound provides pictures of an embryo or fetus within a woman's uterus.

A. *properties of ultrasonography*

1. Most ultrasound scanning is noninvasive (no needles or injections) and is usually painless.

2. Ultrasound is widely available, easy-to-use and less expensive than other imaging methods.

3. Ultrasound scanning gives a clear picture of soft tissues that do not show up well on x-ray images.

4. Ultrasound causes no health problems and may be repeated as often as is necessary.

5. Ultrasound is the preferred imaging modality for the diagnosis and monitoring of pregnant women and their unborn babies.

6. Ultrasound has been used to evaluate pregnancy for nearly four decades and there has been no evidence of harm to the patient, embryo or fetus. Nevertheless, ultrasound should be performed only when clinically indicated.

B. *Obstetrical ultrasound is a useful clinical test to*
- Establish the presence of a living embryo/fetus.
- Estimate the age of the pregnancy.
- Diagnose congenital abnormalities of the fetus.
- Evaluate the position of the fetus.
- Evaluate the position of the placenta.
- Determine if there are multiple pregnancies.
- Determine the amount of amniotic fluid around the baby.
- Check for opening or shortening of the cervix or mouth of the womb.
- Assess fetal growth.

C. *Limitations of Obstetrical Ultrasound Imaging*

Obstetric ultrasound cannot identify all fetal abnormalities. Consequently, when there are clinical or laboratory suspicions for a possible abnormality, a pregnant woman may have to undergo nonradiologic testing such as amniocentesis (the evaluation of fluid taken from the sac surrounding the fetus) or chorionic villus sampling (evaluation of placental tissue) to determine the health of the fetus, or she may be referred by her primary care provider to a gerontologist (an obstetrician specializing in high-risk pregnancies).

III. ELEMENTS OF BIOMEDICAL IMAGE PROCESSING

A. *Image Enhancement*

Enhancement algorithms are used to reduce image noise and increase the contrast of structure of interest. when in images the distinction between normal and abnormal tissue is occur then accurate interpretation may become difficult if noise level are relatively high. Enhancement improve the quality of image and facilitates diagnosis. Enhancement techniques are generally used to provide a clear image for a human observer.

Image enhancement is especially relevant in mammography where the contrast between the soft tissues. These approaches all use a reversible wavelet decomposition, which may be redundant or not, and perform the enhancement by selective modification (Amplification) of certain wavelet coefficients prior to reconstruction. When the weighting scheme is linear, this approach can be interpreted as a multiscale version of traditional unsharp masking.

B. *Image segmentation*

Segmentation is the stage where significant commitment is made during automated analysis by delineating structures of interest and discriminating them from background tissues. Segmentation algorithm operate on the intensity texture variations of the image using technique that include thresholding, region growing and pattern recognition technique such as neural network. Image segmentation is a useful tool in many realms including industry, health care, astronomy, and various other fields. Segmentation in concept is a very simple idea. Simply looking at an image, one can tell what regions are contained in a picture. This paper discusses two different region determination techniques: one that focuses on edge detection as its main determination characteristic and another that uses region growing to locate separate areas of the image.

IV. PROBLEM DESCRIPTION

An image may be defined as a two dimensional function $f(x, y)$, where x and y are spatial (plane) coordinates, and the amplitude of f at any pair of co-ordinates (x, y) is called the intensity or grey level of the image at that point [5]. Data sets collected by image sensor are generally contaminated by noise. The region of interest in the image can be degraded by the impact of imperfect instrument, the problem with data acquisition process and interfering natural phenomena. Therefore the original image may not be suitable for applying image processing techniques and analysis. Thus image enhancement technique is often necessary and should be taken as the first and foremost step before image is analysed. Another common form of noise is data dropout noise generally referred to as speckle noise. This noise is, in fact, caused by errors in data transmission [13, 14]. The corrupted pixels are either set to the maximum value, which is something like a snow in image or have single bits flipped over.

A. *Image Data Independent Noise*

It is described by an additive noise model, where the recorded image, $i(m, n)$ is the sum of the true image $t(m, n)$ and the noise $n(m, n)$[5, 8, 9]

$$i(m,n)=t(m,n)+n(m,n)$$

The noise $n(m, n)$ is often zero-mean and described by its variance. In fact, the impact of the noise on the image is often described by the SNR [6], which is given by

$$SNR = \sqrt{\frac{\sigma_t^2}{\sigma_n^2} - 1}$$

Where, $\sigma_t^2$ and $\sigma_n^2$ are the variances of the true image and the recorded image respectively. In many cases, additive noise is evenly distributed over the frequency domain (white noise), whereas an image contains mostly low frequency information. Therefore, such a noise is dominant for high frequencies and is generally referred as Gaussian noise and it is observed in natural images [10, 11].

*B. Image Data Dependent Noise*

Data-dependent noise (e.g. arising when monochromatic radiation is scattered from a surface whose roughness is of the order of a wavelength, causing wave interference which results in image speckle), it is possible to model noise with a multiplicative, or non-linear, model. These models are mathematically more complicated; hence, if possible, the noise is assumed to be data independent.

*1) Speckle Noise*

Another common form of noise is data dropout noise generally referred to as speckle noise. This noise is, in fact, caused by errors in data transmission [13, 14]. The corrupted pixels are either set to the maximum value, which is something like a snow in image or have single bits flipped over. This kind of noise affects the ultrasound images [14]. Speckle noise has the characteristic of multiplicative noise [15]. Speckle noise

follows a gamma distribution and is given as

$$f(g) = \left[\frac{g^{\alpha-1} e^{g/a}}{(\alpha-1)! a^\alpha}\right]$$

*2) Salt and Pepper Noise*

This type of noise is also caused by errors in data transmission and is a special case of data dropout noise when in some cases single, single pixels are set alternatively to zero or to the maximum value, giving the image a salt and pepper like appearance [16]. Unaffected pixels always remain unchanged. The noise is usually quantified by the percentage of pixels which are corrupted. It is found in mammogram images [17]. It probability distribution of function is in [8].

V. SPATIAL FILTER

The primary objective of the image enhancement is to adjust the digital image so that the resultant image is more suitable than the original image for a specific application [5, 8, and 9]. There are many image enhancement techniques. They can be categorized into two general categories. The first category is based on the direct manipulation of pixels in an image, for instance: image negative, low pass filter (smoothing), and high pass filter (sharpening). Second category is based on the position manipulation of pixels of an image, for instance image scaling. In the first category, the image processing function in a spatial domain can be expressed as

$$g(x, y) = T(f(x, y))$$

Where, T is the transformation function, f (x, y) is the pixel value of input image, and g(x, y) is the pixel value of the processed image [5, 8, 9]. The median, mean, high pass, low pass filtering techniques have been applied for denoising the different images [5, 9].

*A. Max Filter*

The max filter plays a key role in low level image processing and vision. It is identical to the mathematical morphological operation: dilation [19]. The brightest pixel gray level values are identified by this filter. It has been applied by many researchers to remove pepper noise. Though it removes the pepper noise it also removes the block pixel in the border [5]. This filter has not yet applied to remove the speckle in the ultrasound medical image. Hence it is proposed for speckle noise removal from the ultrasound medical image. it is expressed as:

$$f(x, y) = \max\{g(s, t)\}$$

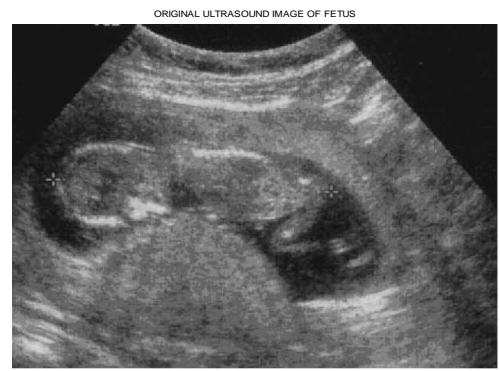

Figure 1. *original image*

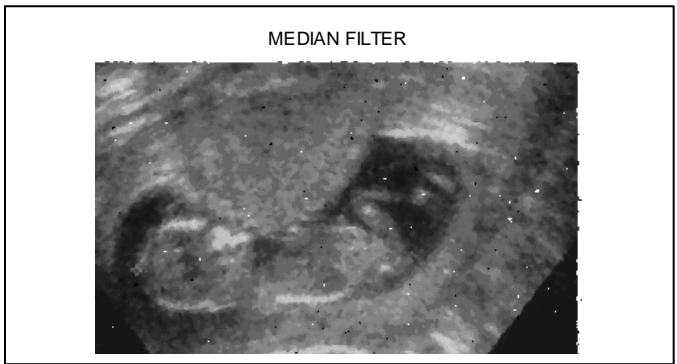

Figure 2. Median filter output.

It reduces the intensity variation between adjacent pixels. Implementation of this method for smoothing images is easy and also reducing the amount of intensity variation between one pixel and the next. The result of this filter is the max selection processing in the sub image area.

### B. Min Filter

The min filter plays a significant role in image processing and vision. It is equivalent to mathematical morphological operation: erosion [19]. It recognizes the darkest pixels gray value and retains it by performing min operation. This filter was proposed for removing the salt noise from the image by researchers. Salt noise has very high values in images. The operation of this filter can be expressed as:

$$f(x,y) = \min\{g(s,t)\}$$

It removes noise better than max filter but it removed some white points around the border of the region of the interest [5]. In this filter each output pixel value can be calculated by selecting minimum gray level value of the chosen window.

### C. Standard Deviation Filter

Normally the interpretations of the images are quite difficult, since the backscatter causes the unwanted noise. The standard deviation was proposed to remove the noise in radar satellite images [2]. This filter has not proposed to remove the speckle noise from the ultrasound medical images to the best of our knowledge. The standard deviation filter [5] calculates the standard deviation for each group of pixels in the sub image, and assigns this pixel in the output image. By using a standard deviation filter, we are able to recognize some patterns. It is expressed as

$$f(x,y) = \sqrt{\frac{1}{n^2} \sum_{r=1}^{n} \sum_{c=1}^{n} \left(x_{rc} - x^{-}\right)^2}$$

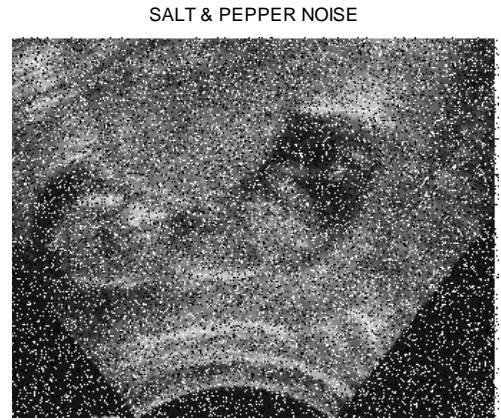

Figure 3. Salp and Pepper Noice output

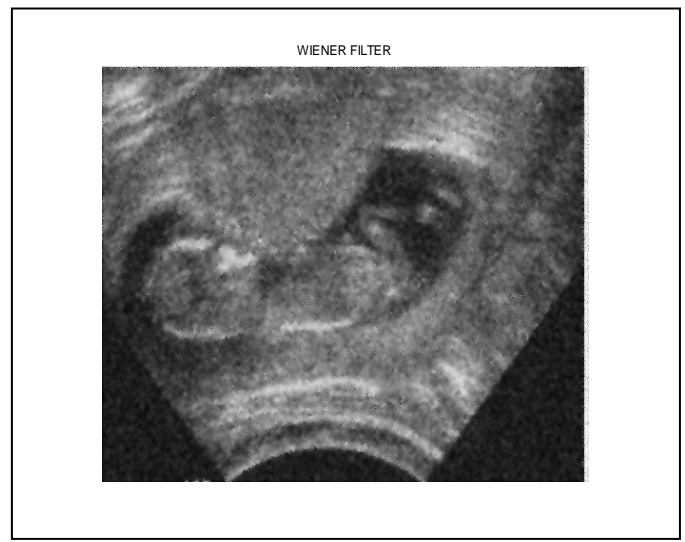

Figure 4. filtered image

Where, n x n is the total number of pixels in a sub image, *the indices of the sub image,* is the value of the pixel at row r and column c in the sub image and *x* is the mean of pixel values in the window. It measures of heterogeneity in the sub image at centered over each pixel. Standard deviation filter is applied to detect the changes in sub images [5]. A small mask was used for the filter in order to obtain sharp edges. A size of 3x3 pixels was supposed to be sufficient. The filter generates a new image based on the value of the standard deviation.

## VI. STATICAL MEASUREMENT

Table; 1 formulas applies for Statistical Measurement

| Statistical Measurement | Formula |
|---|---|
| MSE | $MSE = \dfrac{\sum (f(i,j) - F(i,j))^2}{MN}$ |
| RMSE | $RMSE = \sqrt{\dfrac{\sum (f(i,j) - F(i,j))^2}{MN}}$ |
| SNR | $SNR = 10 \log_{10} \dfrac{\sigma^2}{\sigma_e^2}$ |
| PSNR | $PSNR = 20 \log_{10} \dfrac{255^2}{RMSE}$ |

## VII. COMPUTATIONAL RESULT

Table; 2 Comparative analysis.

| S.No | FILTERING METHOD | RMSE | SNR | PSNR |
|---|---|---|---|---|
| 1. | Median filter | 28.12 | 2.12 | 11.27 |
| 2. | Variance filter | 68.6 | 1.02 | 10.42 |
| 3. | Correlation filter | 80.40 | 2.32 | 5.13 |
| 4. | Mean filter | 102.94 | 9.02 | 6.93 |

## VIII. CONCLUSION

The performance of noise removing algorithms is measured using quantitative performance measures such as PSNR, SNR, and RMSE as well as in term of visual quality of the images. Many of the methods fail to remove speckle noise present in the ultrasound medical image, since the information about the variance of the noise may not be identified by the methods. Performance of all algorithms is tested with ultrasound image regard to fetus.

## REFERENCES

[1] Yanong Zhu, Stuart Williams, Reyer Zwiggelaar, Computer Technology in Detection and Staging of Prostate Carcinoma: A review, Medical image Analysis 10(2006),pp 178-199.

[2] J. G. Abbott and F. L. Thurstone, "Acoustic speckle: Theory and experimental analysis," UltrasonicImag., vol. 1, pp. 303-324, 1979.

[3] D Hykes, W.Hedrick and D.Starchman, Ultrasound Physics and Instrumentation, Churchill New,1985

[4] R.C. Gonzalez and R.E. Woods: 'Digital Image Processing', Addison-Wesley Publishing company, 1992.

[5] ] Image Processing Fundamentals – Statistics, "Signal to Noise Ratio", 2001.

[6] R.C. Gonzalez and R.E. Woods: 'Digital Image Processing', Addison-Wesley Publishing company, 1992.

[7] A.K. Jain, fundamental of digital image processing. Englewood cliffs, NJ Prentice-Hall, 1989.

[8] ] Prostate Carcinoma: A review, Medical image Analysis 10(2006),pp 178-199,2006.

[9] H. GUO, J E Odegard, M.Lang, R.A.Gopinath,I.W.Selesnick, and C.S. Burrus, "Wavelet based Speckle reduction with application to SAR based ATD/R", First Int'I Conf. on image processing , vol.1, pp. 75-79,Nov 1994.

[10] Z. Wang and D. Hang, "Progressive Switching Median Filter for the Removal of Impulse Noise from Highly Corrupted Images ," IEEE Trans. on Circuits and Systems-II: Analog and Digital Signal processing, vol. 46, no. 1, pp.

AUTHORS PROFILE

**V..Rawat** received her B.E in Electrical Engineering from Rajiv Gandhi Technological university ,Bhopal and Master Degree in Instrumentation from Devi Ahiliyabai University,Indore and her field of interest is Instrumentation ,Biomedical Engineering and Image proceesing. She is a member of Biomedical society of India.

**Alok jain** received his B.E in Electronics & instrumentationl Engineering from Samrat Ashok Technological institute ,Vidisha in 1988, and Master Degree in Computer Science from Roorkee in 1992,.he obtained his Ph.D degree from Thapar institute of engineering and technology,patiala in 2006. His field of interest is signal processing ,filter banks,powerelectronics.

**Vibhakar shrimali** is received his B.E. (Electronics & Comm. Engg.) in 1988 from MBM Engg.College Jodhpur (Raj), M.E. (Electronics & Comm.Engg.) in 2003 from Delhi College of Engineering (Delhi) and Ph.D. from IIT, Roorkee in 2009. His field of interest is Electronics & Communication, Medical Electronics, Digital Image Processing and Rehabilitation Engineering.